# TIME EFFICIENT APPROACH TO OFFLINE HAND WRITTEN CHARACTER RECOGNITION USING ASSOCIATIVE MEMORY NET

**Tirtharaj Dash**

B.Tech Final Year Student, Department of Information Technology

National Institute of Science and Technology

Berhampur-761008, India

**Abstract:** In this paper, an efficient Offline Hand Written Character Recognition algorithm is proposed based on Associative Memory Net (AMN). The AMN used in this work is basically auto associative. The implementation is carried out completely in 'C' language. To make the system perform to its best with minimal computation time, a Parallel algorithm is also developed using an API package OpenMP. Characters are mainly English alphabets (Small (26), Capital (26)) collected from system (52) and from different persons (52). The characters collected from system are used to train the AMN and characters collected from different persons are used for testing the recognition ability of the net. The detailed analysis showed that the network recognizes the hand written characters with recognition rate of 72.20% in average case. However, in best case, it recognizes the collected hand written characters with 88.5%. The developed network consumes 3.57 sec (average) in Serial implementation and 1.16 sec (average) in Parallel implementation using OpenMP.

1. Introduction

In the recent years, Hand Written Character Recognition has been a challenging and interesting research area in the field of pattern recognition and image processing (Impedovo et al., 1991; Mori et al., 1992). It contributes mainly to the Human-Computer interaction and improves the interface between the two (Pradeep et al., 2011). Other human cognition methods viz. face, speech, thumb print recognitions are also being great area of research (Imtiaz and Fattah, 2011; Khurana and Singh, 2011; Kurian and Balakriahnan, 2012).

Generally, character recognition can be broadly characterized into two types (i) Offline and (ii) Online. In offline method, the pattern is captured as an image and taken for testing purpose. But in case of online approach, each point of the pattern is a function of time, pressure, slant, strokes etc. Both the methods are best based on their application in the field. Yielding best accuracy with minimal cost of time is a crucial precondition for pattern recognition system. Therefore, hand written character recognition is continuously being a broad area of research.

In this work, an approach for offline character recognition has been proposed using Associative Memory Network (AMN). In fact, to make it time efficient, a parallel algorithm has been developed for the implementation of AMN using OpenMP *(Open Multiprocessing)* (www.openmp.org). AMN is a neural network which can store patterns as memories. When the network is being tested with a key pattern, it corresponds by producing one of the stored patterns, which closely resembles to key pattern. Based on the testing pattern, AMN can be of two types (i) auto-associative memory net or (ii) hereto-associative memory net. Both the networks contains two layers (a) input layer and (b) output layer. In case of auto-associative memory net, the input and target pattern are same (Sivanandam and Deepa, 2011). But, in case of hetero-associative memory net the two patterns are different. This work uses the auto-AMN, as the character to be tested is same as the stored character. The characters considered in this work are English alphabets (both small and capital letters).

This paper is organized as follows. Section 1 presented a general introduction to the character recognition systems and methods. Section 2 gives a brief literature review of some methods proposed for character recognition. Section 3 describes the proposed methodology of this work. Section 4 is a result and discussion section which gives a detailed analysis of the work. The paper is concluded in section 5 with a note to future works.

2. **Literature Review**

Available literatures convey that various algorithms and techniques have been used in order to accomplish the task of character recognition. Some studies are described below. Source of the literature are Google scholar, Scopus and IEEE library.

Neural Network (NN) has been a backend of character classification in most of the methods. This is due to its faster and reliable computation. The methods used in front end could be (a) statistical approaches (b) kernel methods (c) support methods or (d) hybrid of fuzzy logic controllers.

Multilayer Perceptron (MLP) was used for 'Bangla' alphabet recognition by Basu et al., 2005. The accuracy achieved in this work was 86.46% and 75.05% on the samples of training and testing respectively.

Manivannan and Neil, 2010 proposed and demonstrated an optical correlator-neural network architecture for pattern recognition. English alphabet used as patterns for the training and testing process.

(Pal and Singh, 2010) proposed NN based English character recognition system. In this work, MLP with one hidden layer was used. About 500 testing were carried out to test the performance of the design. The best case accuracy obtained in this work was 94%.

(Perwej and Chaturvedi, 2011) worked on English alphabet recognition using NN. In this work, binary pixels of the alphabets were used train the NN. The accuracy achieved was found to be 82.5%.

(Pal et al., 2007) proposed a modified quadratic classifier approach for handwritten numerals of six popular Indian scripts with high level of recognition accuracy.

(Dinesh et al., 2007) used horizontal and vertical strokes and end points as feature for handwritten numerals. This method reported an accuracy rate of 90.5% in best case. However, this method used a thinning method resulting in loss of features.

Yanhua and Chuanjun, 2009 recommended a novel Chinese character recognition algorithm which was based on minimum distance classifier. The algorithm attempted to work with two classes of feature extraction-structure and statistics. The statistic feature decided the primary class and the structure feature used to identify Chinese characters.

A good method of character recognition was proposed by (Huiqin et al, 2011). In this work, they proposed a distribution based algorithm based on image segmentation and distribution of pixels. Deflection Correction method was adopted for flexibility as well as reduction of matching error. This work avoided the burden of extracting the skeleton from the character. The method gave excellent result and was robust.

## 3. Methodology

A step-wise methodology has been proposed which is demonstrated in Figure 1.

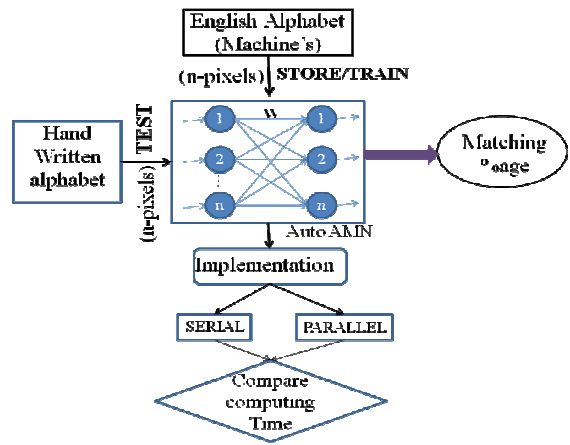

**Figure 1**: Proposed Methodology

**Step-1:** Collection of English alphabets (both small and capital) from (i) system and (ii) persons (Hand Written)

**Step-2:** Extraction of pixels from the characters

**Step-3:** Implementation of auto AMN: (i) Training and (ii) Testing using both (a) serial and (b) parallel algorithms.

**Step-4:** Comparison of results from serial and parallel processing with respect to time of execution

### 3.1 Generation of English alphabets

English alphabets (both small and capital) are designed in the system using *MS Paint version 6.1* in Arial font size-28 (No Bold) in BMP file format. The dimension of the bmp file is 31×39, with bit depth of 4. Some alphabets are given in Figure 2.

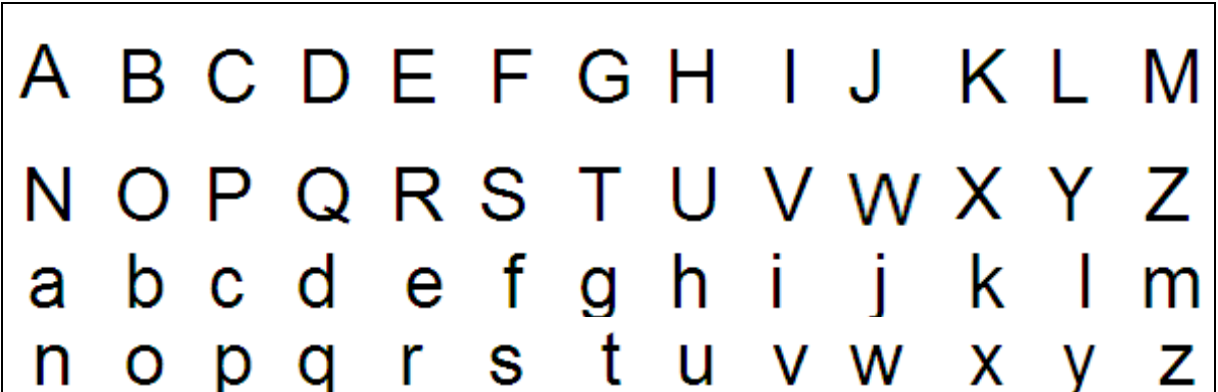

**Figure 2**: English alphabets of the system

Hand written English alphabets are collected, each one from different persons. The characters are given in Figure 3.

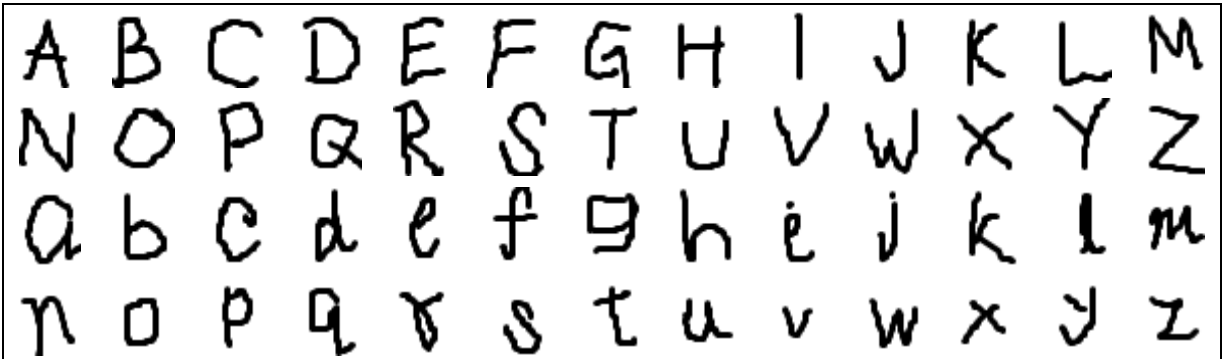

**Figure 3**: English alphabets collected from different persons

### 3.2 Extraction of Pixel from the characters

Pixels are extracted from the character images (bitmap files) using a standard image function of MATLAB version 10. The function is *imread('filename.bmp')*. The function extracts the decimal values associated with each pixel. The pixels are then stored in a text (.txt) file for the experiment purpose.

### 3.3 Auto-associative memory net (Auto-AMN) implementation

#### 3.3.1 Serial algorithm

**INITIALIZE** weight (W) to 0
SET the target pattern as the system's pattern

**INPUT** the Handwritten pattern to the first layer of AMN
**FOR** i=1 to n
**DO**
    **FOR** j=1 to n
    **DO**
        **CALCULATE** the weight as
            $W_{ij}(new) = W_{ij}(old) + INPUT_i \times TARGET_j$
    **END**
**END**
**FOR** i=1 to n
**DO**
    **FOR** j=1 to n
    **DO**
        **CALCULATE** the net input to each output node as,
$$Yin_j = \sum_{i=1}^{n} x_i W_{ij}$$
    **IF**($Yin_j > 0$)
        $Y_j = +1;$
    **ELSE**
        $Y_j = -1;$
    **END**
**END**

### 3.3.2 Parallel algorithm

**INITIALIZE** weight (W) to 0
**SET** the target pattern as the system's pattern
**INPUT** the Handwritten pattern to the first layer of AMN

*#pragma omp paralle shared(W,Yin,chunk,p) private(tid,i,j)*
**DO**

    *#pragma omp for schedule (static, chunk)*
    **FOR** i=1 to n
    **DO**
        **FOR** j=1 to n
        **DO**
            **CALCULATE** the weight as
                $W_{ij}(new) = W_{ij}(old) + INPUT_i \times TARGET_j$
        **END**
    **END**

```
#pragma omp for schedule (static, chunk)
FOR i=1 to n
DO
    FOR j=1 to n
    DO
        CALCULATE the net input to each output node
```

$$Yin_j = \sum_{i=1}^{n} x_i W_{ij}$$

```
        IF(Yin_j>0)
            Y_j=+1;
        ELSE
            Y_j=-1;
        END
    END
END
```

### 3.3.3 System Specification

A computer system having *1 GB RAM* and *Four* processors is used for the complete work. The operating system is *Ubuntu 10.04 (Linux).* However, for auto optimization by the compiler, '–g' tag is used in the compilation command.

### 4. Results and Discussion

The contribution of this work is a detailed analysis on the recognition accuracy for all the handwritten English alphabets (Total 52). Time of computation has been noted for both serial and parallel algorithm to compare the decision making speed.

### 4.1 Recognition accuracy

Table 1 shows a result of the testing the developed AMN for a set of hand written characters. It should be noted that the network is trained with the machine's alphabets and tested with the hand written alphabet. However, for reliability issue, a hand written character is checked for 5 times and the matching percentage is the average of the 5 results.

**Table 1:** Recognition accuracy of AMN for offline Hand written character recognition

| System's Alphabet | Hand Written Alphabet for which highest match is achieved | Recognition Accuracy (%) | Time of Computation (sec.) | |
|---|---|---|---|---|
| | | | Serial | Parallel |
| A | A | 66.56 | 3.00 | 0.99 |
| B | B | 56.80 | 3.11 | 0.98 |
| C | C | 67.19 | 2.98 | 0.74 |
| D | O | 60.88 | 4.60 | 1.55 |
| E | F | 62.85 | 3.77 | 2.01 |
| F | F | 68.34 | 4.65 | 2.32 |
| G | G | 58.78 | 3.12 | 1.87 |
| H | H | 70.73 | 3.67 | 0.99 |
| I | I | 85.74 | 4.01 | 1.03 |
| J | J | 86.28 | 4.32 | 1.93 |
| K | K | 64.85 | 4.17 | 1.05 |
| L | L | 85.16 | 3.04 | 0.87 |
| M | H | 70.38 | 4.12 | 1.21 |
| N | H | 71.13 | 4.61 | 1.35 |
| O | O | 64.17 | 3.02 | 0.76 |
| P | P | 67.74 | 4.00 | 1.04 |
| Q | O | 61.42 | 2.98 | 0.83 |
| R | R | 61.16 | 3.41 | 0.99 |
| S | S | 63.20 | 4.04 | 1.12 |
| T | T | 80.18 | 2.87 | 0.77 |
| U | U | 71.46 | 2.76 | 0.76 |
| V | V | 73.69 | 2.78 | 0.87 |
| W | U | 73.13 | 3.05 | 1.00 |
| X | X | 73.39 | 3.83 | 1.43 |
| Y | Y | 76.67 | 4.05 | 1.45 |
| Z | Z | 64.01 | 4.00 | 1.45 |
| a | o | 69.49 | 3.77 | 1.88 |
| b | b | 65.35 | 3.78 | 1.43 |
| c | e | 73.96 | 3.17 | 1.42 |
| d | d | 67.26 | 3.77 | 1.54 |
| e | e | 67.47 | 3.18 | 1.31 |
| f | p | 73.54 | 3.17 | 1.22 |
| g | y | 69.36 | 3.18 | 1.03 |
| h | b | 70.52 | 4.01 | 1.12 |
| i | i | 83.62 | 3.19 | 1.27 |
| j | j | 88.50 | 4.01 | 1.23 |
| k | K | 69.44 | 3.66 | 1.44 |
| l | l | 83.60 | 3.75 | 1.03 |
| m | m | 62.62 | 2.76 | 1.05 |
| n | n | 72.97 | 2.89 | 0.75 |
| o | o | 70.03 | 2.75 | 0.76 |
| p | p | 68.61 | 3.78 | 0.94 |
| q | q | 65.45 | 3.89 | 0.94 |
| r | r | 82.90 | 4.05 | 1.11 |
| s | s | 68.66 | 3.00 | 0.77 |
| t | t | 79.92 | 2.01 | 0.52 |
| u | u | 78.54 | 4.89 | 1.33 |
| v | v | 82.52 | 4.02 | 1.04 |
| w | w | 67.16 | 3.96 | 1.21 |
| x | x | 78.55 | 2.95 | 0.76 |
| y | y | 78.27 | 3.81 | 0.98 |
| z | z | 70.08 | 3.99 | 1.23 |

Table-1 can be viewed as a detailed analysis of performance of the developed auto AMN for offline hand written English alphabet recognition. The network recognizes the handwritten character 'j' with highest matching of *88.50%*. However, the network doesn't recognize some alphabets like 'D', 'E', 'M', 'N', 'Q', 'W', 'a', 'c', 'f', 'g', 'h', 'j' and 'k' and these alphabets are recognized as 'O', 'F', 'H','H', 'O', 'U', 'o', 'e', 'p', 'y', 'b', 'I' and 'K' respectively with some matching error.

### 4.2 Level of matching of each alphabet

A plot has been given in Figure 4 to view the level up to which each English alphabet is matched by the AMN. The alphabets which are not recognized are awarded with 0% matching.

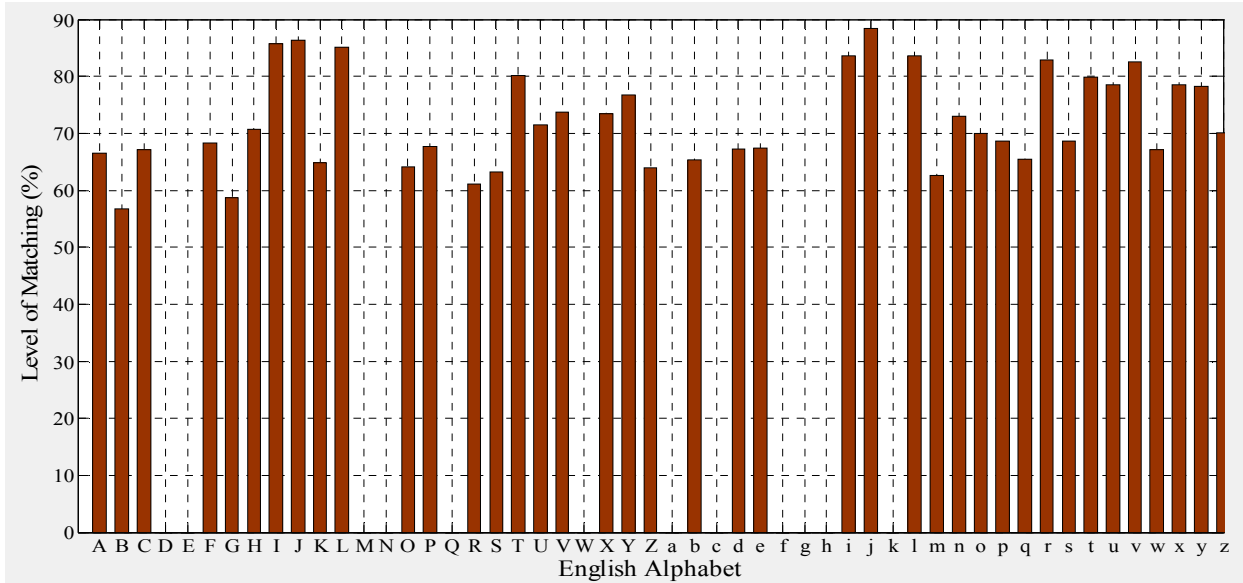

**Figure 4** This plot shows level of matching of each alphabet

**4.3 Time Efficiency**

As it is already mentioned that the network is developed with two algorithms, (i) serial and (ii) parallel; it will be a good idea to check the timing variation in both the cases. A plot given in Figure 5, shows speed up after achieved after the execution of parallel algorithm.

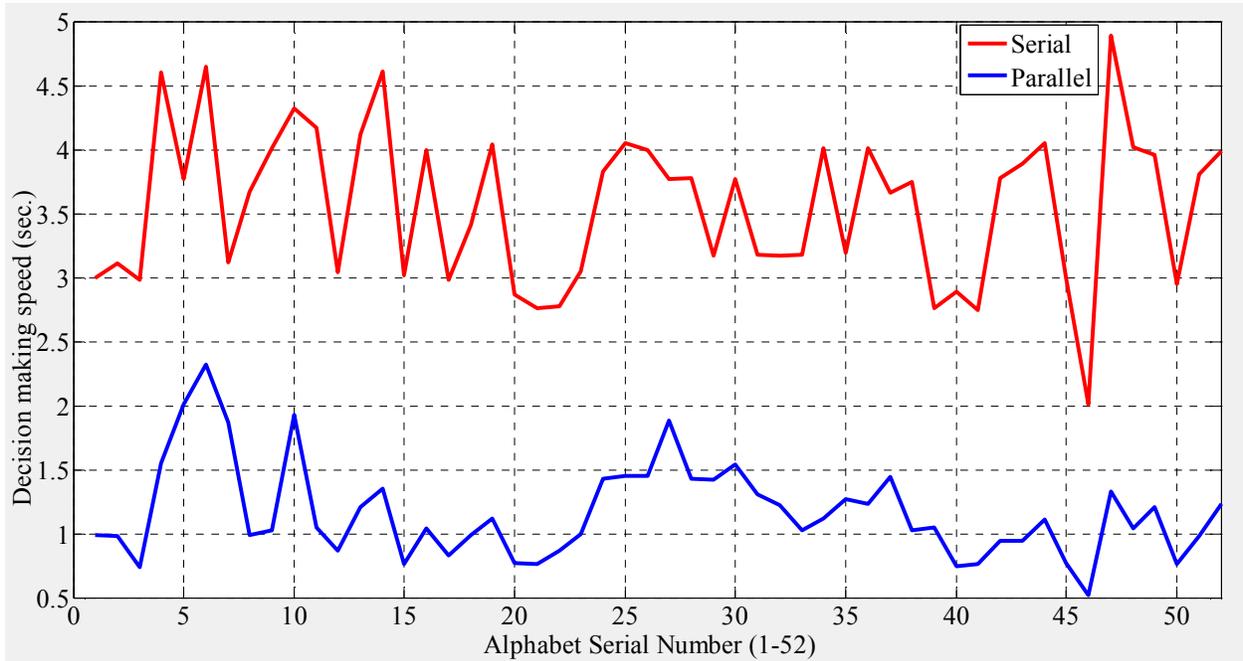

**Figure 5** Decision making speed by Serial and Parallel algorithm

### 5. Conclusion

In this paper, an offline English character recognition system has been proposed. The system is developed using auto Associative Memory Net. To make the developed system faster and reliable, a parallel algorithm has been developed and tested successfully. Experimental study showed that, the system recognizes characters with average recognition rate of **72.20%**. Character *'j'* is recognized with highest accuracy rate of *88.5%*. The average time required by the serial algorithm to recognize a character is *3.57 sec* where as the parallel algorithm takes only *1.16 sec* on an average. However, automatic checking of a sequence of character by the network will play a great role in the world of character recognition. The author is currently working on this issue.